\documentclass[conference]{IEEEtran}
\IEEEoverridecommandlockouts
\usepackage{cite}
\usepackage{amsmath,amssymb,amsfonts}
\usepackage{graphicx}
\usepackage{textcomp}
\usepackage{xcolor}
\usepackage{booktabs}
\usepackage[caption=false,font=footnotesize]{subfig}
\usepackage[spaces,hyphens]{xurl} 
\usepackage{multirow}
\usepackage{multicol}
\usepackage{algorithm}
\usepackage{algorithmicx}
\usepackage{amsmath}
\algnewcommand{\LeftComment}[1]{\Statex \(\triangleright\) #1}
\usepackage[noend]{algpseudocode}
\usepackage{comment}
\usepackage{changepage,threeparttable} 

\newcommand{\etal}{{\textit{et al.}}}

\def\BibTeX{{\rm B\kern-.05em{\sc i\kern-.025em b}\kern-.08em
    T\kern-.1667em\lower.7ex\hbox{E}\kern-.125emX}}
\begin{document}
\title{Minimally Supervised Topological Projections of Self-Organizing Maps for Phase of Flight Identification\\}



\author{\IEEEauthorblockN{Zimeng Lyu}
\IEEEauthorblockA{
\textit{Rochester Institute of Technology}\\
Rochester, NY, USA \\
zimenglyu@mail.rit.edu}
\and
\IEEEauthorblockN{Pujan Thapa}
\IEEEauthorblockA{
\textit{Rochester Institute of Technology}\\
Rochester, NY, USA \\
pt6757@rit.edu}
\and 
\IEEEauthorblockN{Travis Desell}
\IEEEauthorblockA{
\textit{Rochester Institute of Technology}\\
Rochester, NY, USA \\
tjdvse@rit.edu}
}

\maketitle

\begin{abstract}
Identifying phases of flight is important in the field of general aviation, as knowing which phase of flight data is collected from aircraft flight data recorders can aid in the more effective detection of safety or hazardous events. General aviation flight data for phase of flight identification is usually per-second data, comes on a large scale, and is class imbalanced. It is expensive to manually label the data and training classification models usually faces class imbalance problems. This work investigates the use of a novel method for minimally supervised self-organizing maps (MS-SOMs) which utilize nearest neighbor majority votes in the SOM U-matrix for class estimation. Results show that the proposed method can reach or exceed a naive SOM approach which utilized a full data file of labeled data, with only 30 labeled datapoints per class. Additionally, the minimally supervised SOM is significantly more robust to the class imbalance of the phase of flight data. These results highlight how little data is required for effective phase of flight identification.

\end{abstract}


\section{Introduction}

Data classification is widely used in many different domains, such as health care\cite{akhtar2023big}, environmental science\cite{wang2021seppcnet}, marketing\cite{saura2021using}, manufacture\cite{li2022review}, and finance\cite{dolphin2022multimodal}. In real-world applications, data is noisy, potentially large-scale, and usually unlabeled with potential class imbalance problems. Labeling real-world, high-dimensional, large-scale data can be very costly and time-consuming, especially when expert domain knowledge is required. In many cases, it can require human experts to manually label the data or to have data sent to labs for analysis, etc. Developing minimally supervised learning methods, especially for large-scale class imbalanced data can provide significant benefits in many application areas.

One such area is General Aviation (GA), which is all civilian flying except for scheduled commercial passenger airline services. Safety is one of the primary concerns in the general aviation industry as accidents are much more common in GA flights than in commercial flights. According to the General Aviation Accident Dashboard published by the National Transportation Safety Board (NTSB), a total of 1437 GAaccidents occurred in the year 2021 where 946 were non-fatal and 210 were fatal \cite{aviation_dashboard} -- rates significantly higher than commercial aviation. To work towards safety, many operators within the GA community perform Flight Data Monitoring (FDM) to improve the safety of the flight and aircraft~\cite{kuo2017search, liu2022design}.

The identification of what flight data corresponds to what phase of flight can aid in improving flight safety by leading to new and more efficient detection and mitigation strategies.
However identifying of phases of flight in GA is more challenging than commercial flights because GA flights have more altitude changes and are often flown manually, whereas most commercial flights are autopilot~\cite{goblet2015identifying}. Unfortunately, in the field of GA, rules for determining the phase of flight are not granular enough, and available flight sensors are limited, making automating the phase of flight identification a challenging task.

\begin{figure}[t]
    \centering
    \includegraphics[width=\columnwidth]{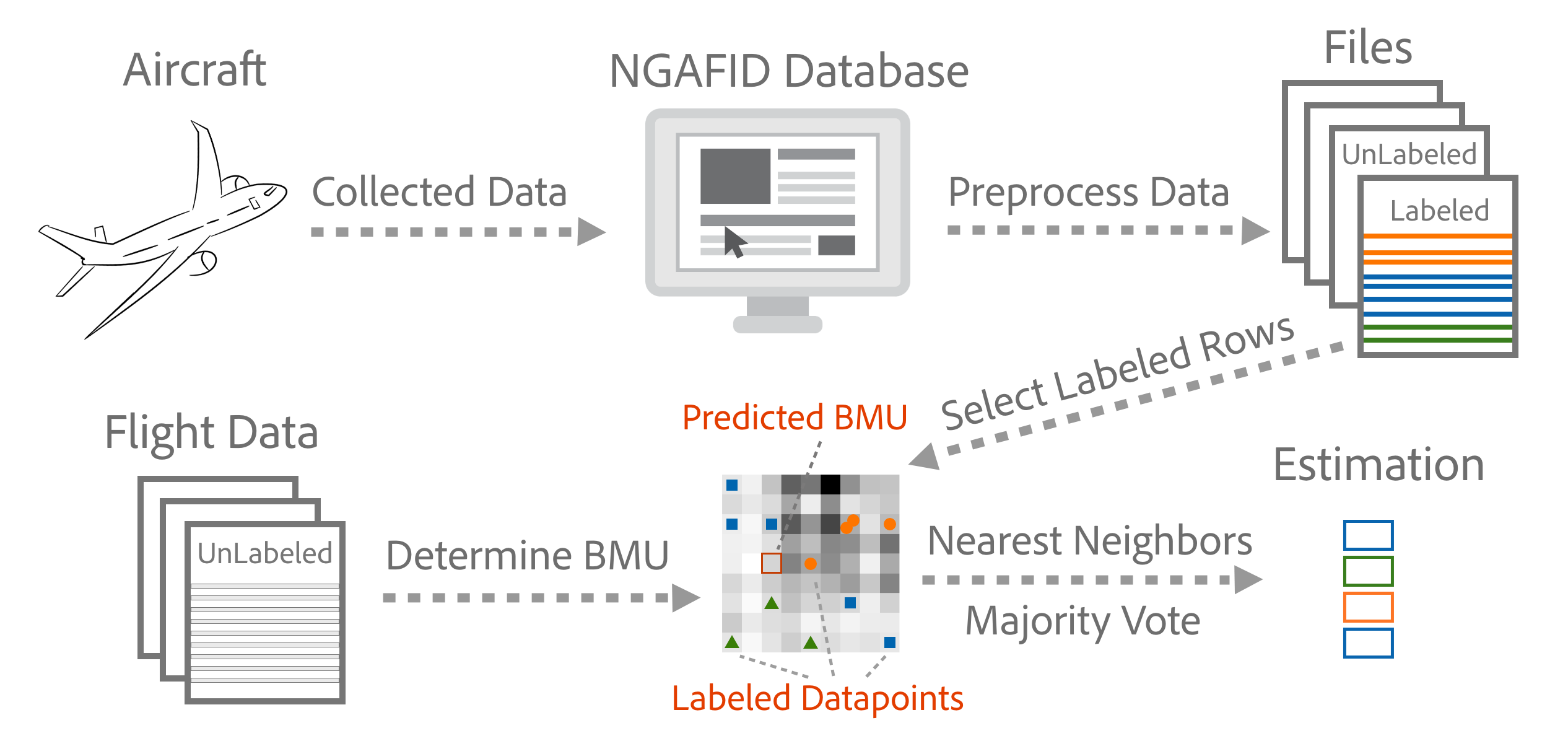}
    \caption{Flow diagram for the data ingestion, selection, preprocessing, SOM training, and phase of flight estimation.}
    \label{fig:som}
\end{figure}

Previous works have been done in this area to identify the phases of the flight. Goblet et al. \cite{goblet_phases} worked on identifying the detailed phases of flight by using the definitions given by the NTSB~\cite{flight_phases_ntsb} and found that a logical regression improved the accuracy in determining the unknown flight phases. Bektas~\cite{bektas_2023} and Faure et al.~\cite{faure_som} used the idea of self-organizing maps (SOMs) and their dimensionality reduction ability~\cite{kohonen} to distinguish between the flight phases. They utilized univariate signals to distinguish the flight phases, as opposed to this and other works that utilize multivariate time series flight data. Zhang et al.~\cite{reliable_fuzzy_simulation} studied different approaches in labeling the phases of flight, however, they utilized data from the Automatic Dependent Surveillance-Broadcast (ADS-B) which did not provide ground truths for labeling. To overcome lack of labeled data, they instead used synthetic data for validation of their model, whereas in our work data consists of real flight data with manually labeled phases of flight, which we have made public to benefit the community.

\begin{algorithm*}[!t]
    \begin{algorithmic}[1]
        \caption{MS-SOM Topological Projection for Classification}
        \label{alg:som}
        \footnotesize
        \Function{SOM}{$X\textrm{-}{unlabeled}$, $X\textrm{-}{labeled}$, $Y\textrm{-}{labeled}$}
            \State $som = SOMClustering.fit(X\textrm{-}{unlabeled})$
            \Comment{\emph{Cluster unlabeled data with SOM}}
            \State $U\textrm{-}Matrix = som.getU\textrm{-}Matrix()$
            
            \State $DistanceGraph$ = $DijkstraGraph(UMatrix)$
            \Comment{\emph{Build Dijkstra Shortest Paths Distance Graph}}
            
            \State $PairwiseDistance = DistanceGraph.getPairwiseDistance()$
            \Comment{\emph{Shortest distance between any two nodes}}
            
            \State $Labeled\textrm{-}BMUs = som.transform(X\textrm{-}{labeled})$
            \Comment{\emph{Fit labeled data on trained SOM}}
            
            \State $ClosestNeighbors = findClosestNeighbor(PairwiseDistance, Labeled\textrm{-}BMUs, N)$
            
            \Comment{\emph{Find N closest neighbors and their pairwise distances for each node}}

            \State $EstimatedValues = MakeMajorityVoteTable(ClosestNeighbors, Y\textrm{-}{labeled}, N)$
            \Comment{\emph{Calculate estimated values for each node}}

            \For{$x$ in $X\textrm{-}{unlabeled}$}
                \State $x\textrm{-}BMU = som.transform(x)$
                \Comment{Get unknown data point's BMU}
                \State $y\textrm{-}{estimated} = EstimatedValues[x\textrm{-}BMU]$
                \Comment{Look up its estimated values}
            \EndFor
        \EndFunction

        \Function{MakeMajorityVoteTable}{$ClosestNeighbors$, $Y\textrm{-}{labeled}$, $N$}
            \For{$x$ in $X\textrm{-}{unlabeled}$}
                \For{$n$ in $N$}
                    \State $counter = Count(classes)$
                    \State $MajorityVote = counter.MostCommon()$
                \EndFor
                \State $Estimation[x] = MajorityVote$
            \EndFor
            \State $return Estimation$
        \EndFunction

    \end{algorithmic}
\end{algorithm*}

In addition to the importance of identifying the phase of flight, understanding the flight data is crucial. Flight data is usually class imbalanced because different flight phases take different amounts of time. For example, it only takes a few seconds to a minute for a plane to take off and land, but the time spent while an aircraft is en route to a destination is significantly longer than the other flight phases. Prior work has utilized data augmentation to balance the classes and further improve the model performance~\cite{ali2019imbalance, jiang2020data}. Other researchers have investigated utilizing methods to prevent the model itself from being biased on the majority classes~\cite{tasci2022bias, hong2021unbiased}. Other minimally supervised or semi-supervised methods look into using less labeled data for class-imbalanced applications. Yang~\etal~proposed a pre-trained CNN model for a classifier to detect poor quality images~\cite{yang2022minimally} and Zaheer~\etal~have incorporated neural networks and clustering algorithms for minimally supervised anomaly detection~\cite{zaheer2021cleaning}.

Combining Self Organizing Map with distance-based methods, especially K-nearest neighbors (KNN) for classification~\cite{silva2011som}, anomaly detection~\cite{tian2014anomaly}, dimensionality reduction \cite{wu2010som} have been widely studied. In particular, recent work by Lyu \etal~using topological projections within SOM U-Matrices utilizing minimal labeled data has shown great potential in parameter estimation (regression) for minimally supervised learning on real-world datasets~\cite{lyu2024minimally}. This paper extends this algorithm for classification tasks by applying a majority vote of SOM topological nearest neighbors and applies it to the phase of flight identification.

Experiments are conducted using different amounts of labeled flight phase data to perform topological projections with SOMs trained on unlabeled data, the trained SOMs are tested against a naive SOM classification strategy which requires a large amount of labeled data, as well as the minimally supervised SOM (MS-SOM) method. Results show that the proposed MS-SOM method can reach similar overall accuracy to the naive approach, using only 30 labeled data points per class, vs. a complete file of 4k labeled data points, while performing significantly better on mean per-class accuracy (as the naive approach fails to adequately predict classes with a low number of samples), showing that it is more robust to datasets with class imbalances.



\section{Methodology}
\label{sec:methodology}

\begin{figure}
    \centering
    \includegraphics[width=0.8\columnwidth]{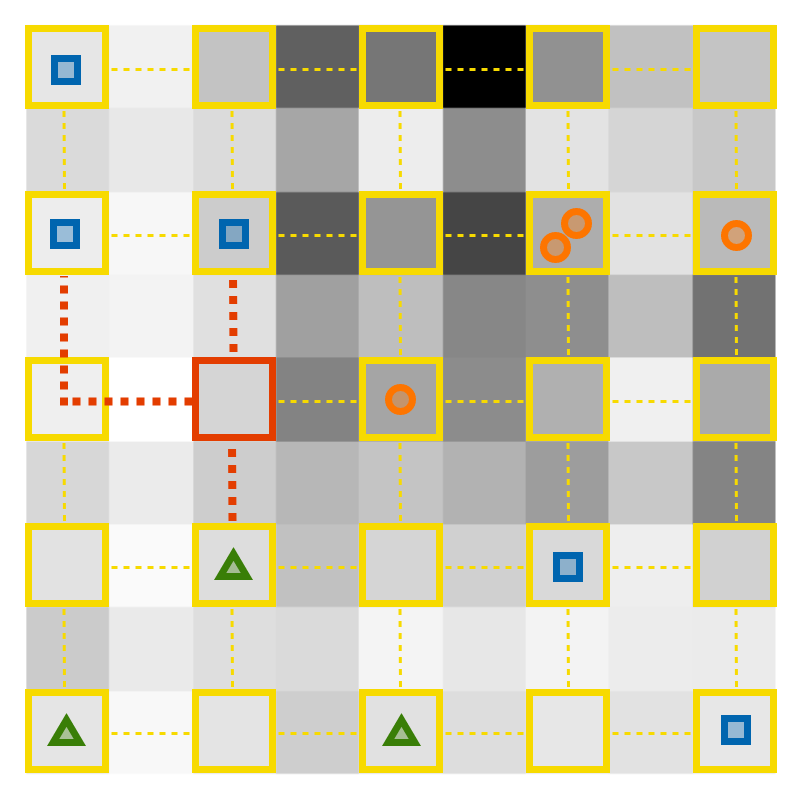}
    \caption{An example topological projection of an unlabeled data
point to its nearest labeled neighbors in a trained SOM topology.}
    \label{fig:umatrix}
\end{figure}

\begin{figure}
  \centering
  \includegraphics[width=\columnwidth]{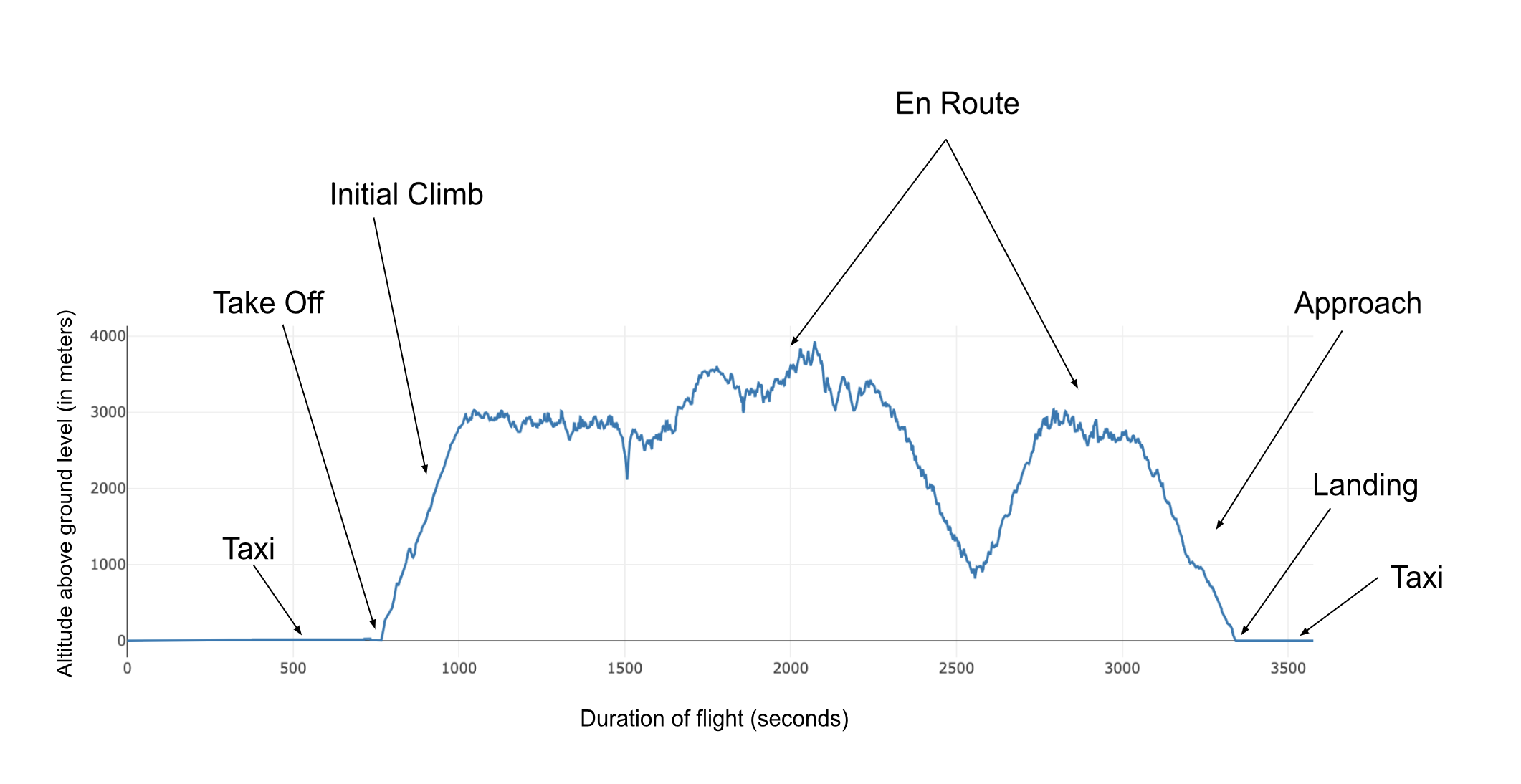}
  \caption{Different phases of a flight from its initial runway taxi phase to its final landing taxi phase.}
  \label{fig:Flight_Phases}
\end{figure}

Self-organizing maps (SOM), also called Kohonen maps~\cite{kohonen}, are an unsupervised machine learning technique used to produce a low dimensional representation of a higher dimensional data set. SOMs cluster high-dimensional data into a two-dimensional grid, which allows these clusters of high-dimensional data to be easily visualized in a 2D plot called a U-Matrix. U-Matrices are typically either a square grid, where each cell has four neighbors or a hexagonal grid where each cell has 6 neighbors. A U-Matrix not only shows the clustering result but also uses a staggered approach where unit cells have connecting cells that are shaded to represent the distance between them. Figure~\ref{fig:umatrix} shows an example of a U-Matrix, where the cells with yellow bounds are actual SOM units, and the cells between the yellow bounds represent connecting cells with darker colors representing more distance between nodes.

\begin{table*}
  \caption{Phases of Flight}
  \label{tab:freq}
  \begin{tabular}{l |p{15cm}l}
    \toprule
    Phase of Flight&Definition\\
    \midrule
    Taxi (TXI) & The aircraft is moving on the aerodrome surface under its own power prior to takeoff or
after landing. \\
    Take off (TOF) & From the application of takeoff power, through rotation and to an altitude of 35 feet above
runway elevation. \\
    Initial Climb (ICL) & From the end of the Takeoff subphase to the first prescribed power reduction, or until
reaching 1,000 feet above runway elevation.\\
    En Route (ENR) & Any level of flight segment after the initial climb until the start of descent to the destination.\\
    Approach (APR) & From 1,000 feet above the runway elevation, to the beginning of the landing flare.\\
    Landing (LDG) & From the beginning of the landing flare until aircraft exits the landing runway, comes to a
stop on the runway. \\
  \bottomrule
\end{tabular}
\end{table*}

\begin{table}
\centering
\caption{Number of Rows Representing Each Phase of Flight in Each Flight File}
\label{tab:row_counts}
\scriptsize
\begin{tabular}{l|cccccc|c}
\toprule
{Flight ID} & { TXI} & {TOF} & { ICL} & { APR} & {LDG} & {ENR} & { Total} \\
\midrule
53430 & 841 & 59 & 483 & 620 & 80 & 2224 & 4307 \\
53433 & 811 & 2 & 131 & 258 & 4 & 1623 & 2829 \\
53434 & 916 & 28 & 472 & 563 & 91 & 1909 & 3979 \\
53435 & 2031 & 15 & 802 & 1085 & 14 & 11705 & 15652 \\
53436 & 1049 & 11 & 172 & 158 & 11 & 9090 & 10491 \\
53438 & 1333 & 16 & 780 & 877 & 54 & 721 & 3781 \\
\midrule
{Total } & 6981 & 131 & 2840 & 3561 & 254 & 27272 & 41039 \\
\bottomrule
\end{tabular}
\end{table}

\subsection{Naive Inference in Self-Organizing Maps}
\label{sec:naive_som_method}

While SOMs are most commonly used as an unsupervised clustering technique, they can also be used to perform inference for classification problems if the data being clustered is labeled. After the SOM has been trained, each sample in the dataset can be assigned to its best matching unit (BMU) which can accumulate the labels of every sample matched to it. Following this, for inference whenever an unlabeled data point is matched to a unit in the SOM, its class can be predicted by using the majority vote of the dataset samples that were mapped to the unit.

\subsection{Topological Projections Using Minimally Supervised Self-Organizing Maps (MS-SOMs)}
\label{sec:ms_som_method}
Recently, SOMs have shown great potential for minimally supervised parameter value estimations for high dimensional real-world data using a topological projection method~\cite{lyu2024minimally}. This work adapts the work by Lyu \etal~to use topological projections to perform classification using a minimal amount of labeled data on SOMs trained on a large amount of unlabeled data. This work is based on the fact that the U-Matrix and the inner representation of a SOM is an undirected graph, where the edges between each SOM unit can be assigned a weight that can be calculated as the distance between those two SOM units (using whatever distance metric the SOM is utilizing during training). Given a trained SOM graph, labeled data points can be assigned to units in the SOM by matching them to the unit with the smallest distance to the labeled data point parameters. After that, inference can be performed on other unlabeled data points by first finding their best matching unit in the SOM and then determining their nearest N units with assigned labeled data points in the graph using Dijkstra's shortest path algorithm (shown in Fig~\ref{fig:umatrix}). A consensus of the class of these N-nearest labeled neighbors can provide the predicted class from the SOM. Algorithm~\ref{alg:som} provides a version of the algorithm presented by Lyu \etal~\cite{lyu2024minimally} updated to perform classification.

\begin{table}
  \caption{Naive SOM Validation and Testing Accuracy}
  \label{tab:som_38}
  \centering
  \begin{tabular}{c|ccccc}
    \toprule
    Size & 5x5  & 10x10 & 15x15 & 20x20 & 25x25\\
    \midrule
     & \multicolumn{5}{c}{Minmax Norm} \\
     \midrule
    Validation & 78.26 & 80.17 & \bf{85.17} & 76.77 & 76.30\\
    Test       & 77.97 & 80.79 & ~\bf{83.35} & 80.92 & 82.70 \\
    \midrule
    & \multicolumn{5}{c}{Standard Norm} \\
    \midrule
    Validation & 79.66 & \bf{80.37} & 75.27 & 66.77 & 74.11\\
    Test     & 68.87 & 71.79 & \bf{81.11} & 79.23 & 73.42  \\
    \midrule
    & \multicolumn{5}{c}{Robust Norm} \\
    \midrule
    Validation & 70.26 & \bf{80.17} & 82.66 & 68.22 & 78.22\\
    Test       & 73.97 & 76.22 & 75.35 & ~\bf{79.22} & 74.69 \\
  \bottomrule
\end{tabular}
\end{table}

\begin{figure}
    \centering
    \setkeys{Gin}{width=0.5\linewidth}
\subfloat[15x15 U-Matrix]{\includegraphics{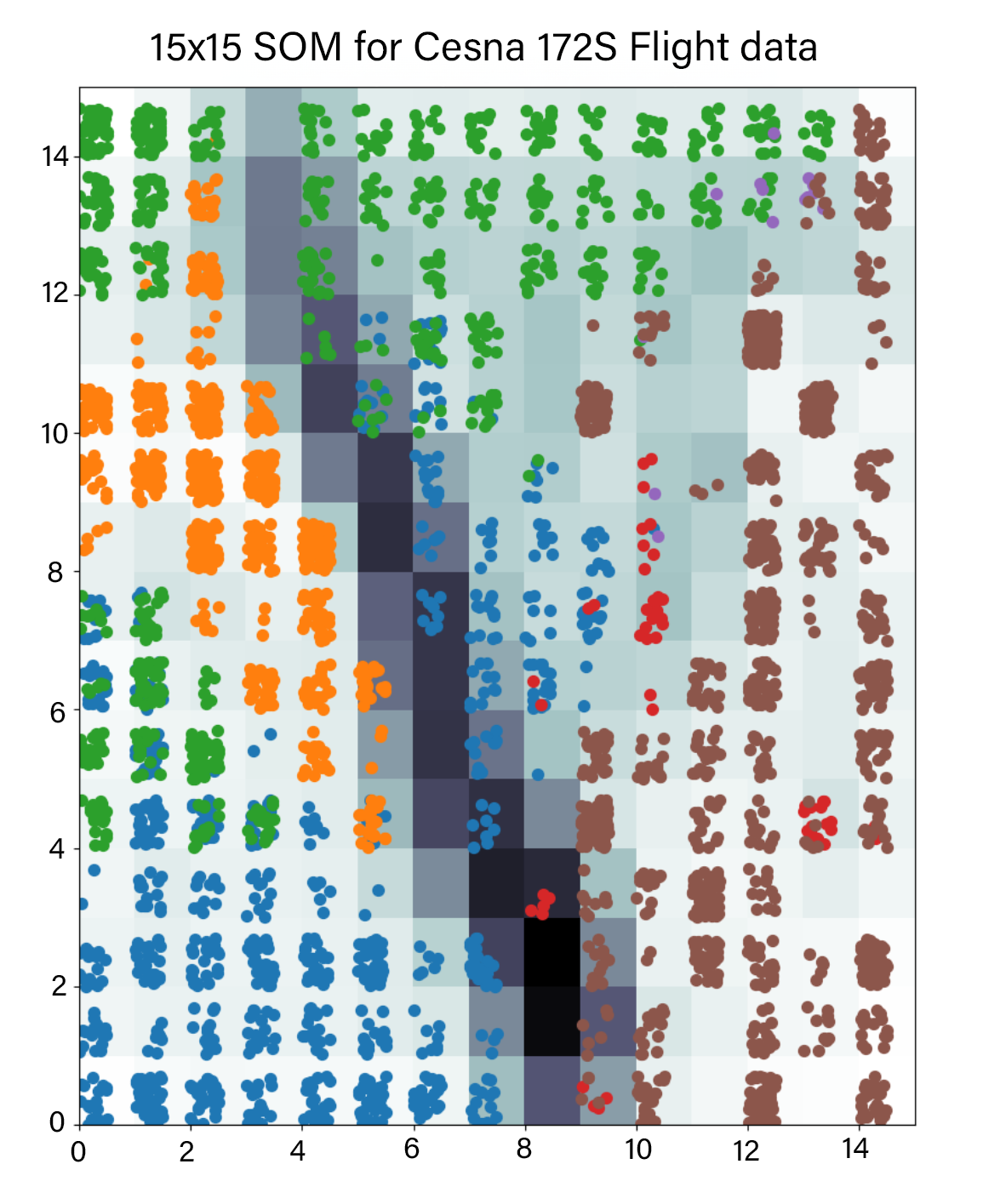}}
\subfloat[Distance Indicator (left) and Label legend (right)]{\includegraphics[height=5cm]{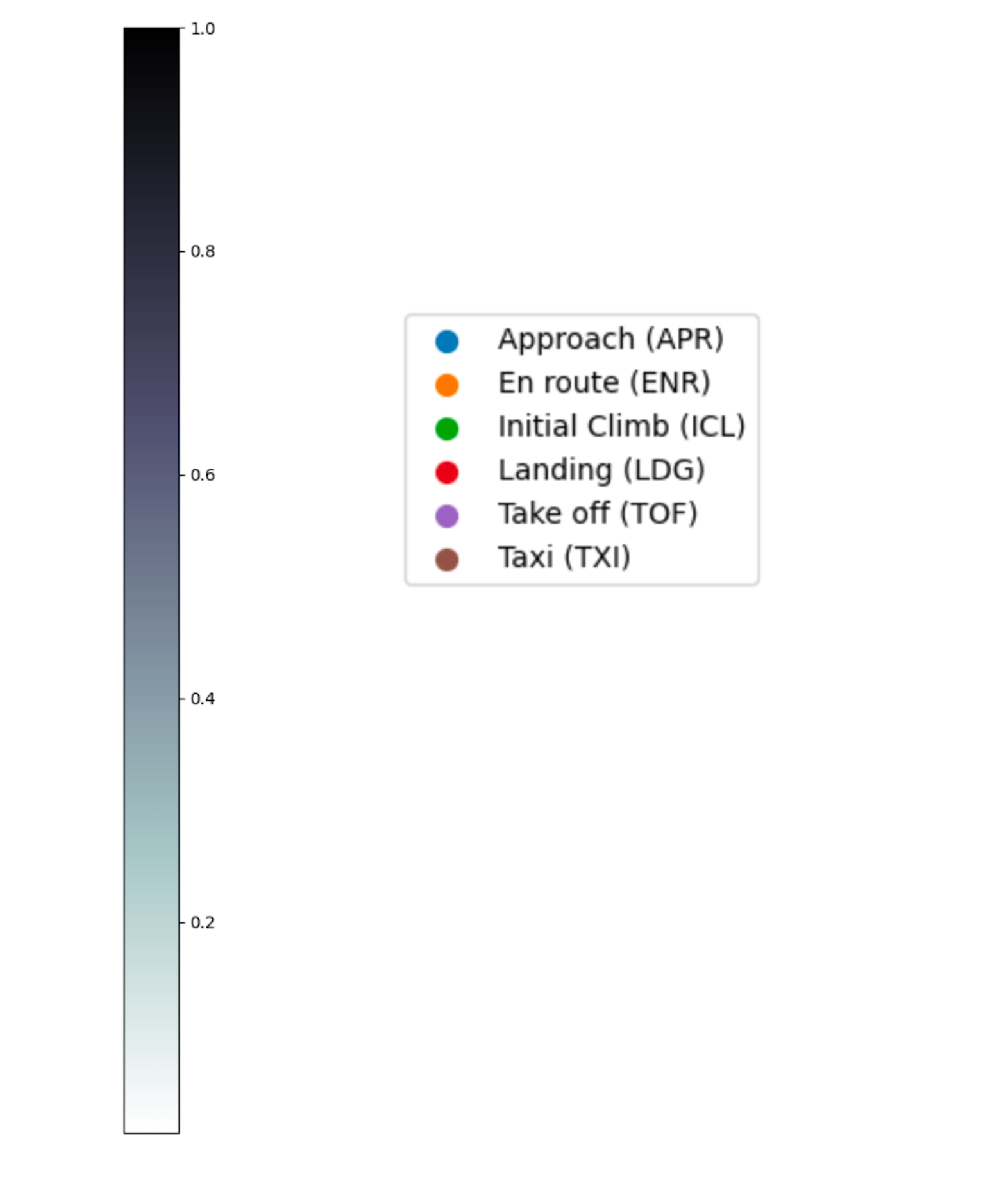}}

\caption{U-matrix for the best found 15x15 SOM trained with Flight 53438 data.}
\label{fig:umatrix-Naive-som}
\end{figure}

\begin{table}
\centering
\caption{SOM Configurations}\label{tab:train_config}
\scriptsize
\begin{tabular}{c|rl}\toprule
Labels Per Class & \multicolumn{2}{c}{Configuration } \\
\midrule
\multirow{2}{*}{5} & SOM Size: & 5x5, 7x7, 10x10\\
                   & Number Neighbors: & 3, 5, 7 \\
\midrule
\multirow{2}{*}{10} & SOM Size: & 5x5, 7x7, 10x10, 15x15 \\
                    & Number Neighbors: & 3, 5, 7, 10 \\
\midrule
\multirow{2}{*}{20, 30, 40} & SOM Size: & 5x5, 7x7, 10x10, 15x15, 20x20, 25x25 \\
                      & Number Neighbors: & 3, 5, 7, 10, 15 \\
                
\bottomrule
\end{tabular}
\end{table}

\begin{table*}[t]\centering
\caption{Results from Inference using Labeled Data Selected From Only Flight 53438}\label{tab:38_test_all}
\scriptsize
\begin{tabular}{c|ccccc|ccccc|ccccc}\toprule
&\multicolumn{5}{c|}{Minmax Norm} &\multicolumn{5}{c|}{Standard Norm} &\multicolumn{5}{c}{Robust Norm} \\\cmidrule{2-16}
\# Neighbors: &3 &5 &7 &10 &15 &3 &5 &7 &10 &15 &3 &5 &7 &10 &15 \\\midrule

SOM Size & \multicolumn{15}{c}{5 Labels Per Class} \\
\midrule
5 x 5 & 25.19 & 22.61 & 25.00 & - & - & 73.42 & 29.70 & 20.32 & - & - & 67.40 & 68.29 & 7.23 & - & -\\ 
7 x 7 & 78.34 & 71.53 & 61.50 & - & - & 76.02 & \bf{81.04} & 74.74 & - & - &67.25 & 67.25 & 67.53 & - & -\\ 
10 x 10 & 80.83 & 75.22 & 61.64 & - & - & 79.77 & 77.57 & 64.48 & - & - &67.53 & 68.72 & 68.46 & - & -\\ 
\midrule
& \multicolumn{15}{c}{10 Labels Per Class} \\
\midrule
5 x 5 & 78.98 & 78.72 & 23.40 & 27.44 & - &  69.50 & 74.98 & 74.10 & 30.60 & - & 67.37 & 68.53 & 68.21 & 68.21 & -\\ 
7 x 7 & 75.70 & 77.22 & 77.30 & 77.13 & - & 73.86 & 76.08 & 78.90 & 77.92 & - & 66.99 & 67.01 & 66.99 & 66.98 & -\\ 
10 x 10 & 80.69 & 79.93 & 79.01 & 73.65 & - & 75.67 & 76.09 & 74.74 & 73.62 & - & 67.14 & 67.14 & 66.89 & 68.24 & -\\ 
15 x 15 & \bf{82.99} & 81.74 & 80.22 & 75.44 & - & 75.20 & 76.88 & 74.49 & 75.44 & - & 52.34 & 68.45 & 67.03 & 66.83 \\ 
\midrule
& \multicolumn{15}{c}{20 Labels Per Class} \\
\midrule
5 x 5 & 74.46 & 73.82 & 73.66 & 70.95 & 23.04 & 79.94 & 73.71 & 73.84 & 76.85 & 79.59 & 68.53 & 68.39 & 68.39 & 68.21 & 7.23 \\ 
7 x 7 & 81.57 & 81.52 & 81.31 & 79.48 & 72.59 & 77.98 & 70.37 & 74.94 & 80.56 & 82.17 & 67.06 & 67.43 & 66.99 & 66.99 & 67.00 \\ 
10 x 10 & 82.16 & 80.59 & 81.41 & 78.45 & 79.78 & 79.11 & 77.27 & 78.98 & 79.87 & 78.43 & 68.61 & 68.56 & 68.52 & 68.28 & 68.28 \\ 
15 x 15 & 80.91 & 80.89 & 81.55 & 81.40 & 79.09 & 79.26 & 77.88 & 78.58 & 80.04 & 78.79 & 68.41 & 68.31 & 68.27 & 68.19 & 67.13 \\ 
20 x 20 & 81.18 & 80.44 & 81.17 & 81.12 & 80.19 & 80.53 & 80.10 & 80.77 & 81.26 & 78.19 & 14.95 & 68.07 & 68.22 & 29.04 & 66.96 \\ 
25 x 25 & 81.91 & 82.47 & 82.31 & \bf{82.49} & 81.40 & 77.10 & 75.75 & 75.39 & 73.82 & 72.91 & 68.70 & 68.49 & 68.41 & 68.59 & 68.39 \\ 
\bottomrule
\end{tabular}
\end{table*}

\begin{table*}[t]\centering
\caption{Results from Inference using Labeled Data Selected Randomly From 3 Flights}\label{tab:343638_test_all}
\scriptsize
\begin{tabular}{c|ccccc|ccccc|ccccc}\toprule
&\multicolumn{5}{c|}{Minmax Norm} &\multicolumn{5}{c|}{Standard Norm} &\multicolumn{5}{c}{Robust Norm} \\\cmidrule{2-16}
\# Neighbors: &3 &5 &7 &10 &15 &3 &5 &7 &10 &15 &3 &5 &7 &10 &15 \\\midrule

SOM Size & \multicolumn{15}{c}{5 Labels Per Class} \\
\midrule
5 x 5 & 39.16 & 38.00 & 37.53 & - & - & 60.90 & 39.50 & 48.04 & - & - & 35.98 & 35.87 & 35.18 & - & -\\ 
7 x 7 & 46.40 & 40.35 & 35.26 & - & - & 69.03 & 63.26 & 62.95 & - & - & 60.62 & 45.85 & 48.52 & - & -\\ 
10 x 10 & 45.20 & 51.29 & 40.37 & - & - & \bf{70.34} & 69.73 & 48.90 & - & - & 65.17 & 57.74 & 65.05 & - & -\\ 
\midrule
& \multicolumn{15}{c}{10 Labels Per Class} \\
\midrule
5 x 5 & 29.63 & 18.00 & 40.79 & 40.00 & - & 67.75 & 66.57 & 48.43 & 47.38 & - & 42.86 & 42.86 & 44.74 & 44.47 & - \\ 
7 x 7 & 66.64 & 55.92 & 43.64 & 46.84 & - & 64.97 & \bf{75.11} & 70.90 & 73.03 & - & 56.34 & 55.53 & 63.38 & 63.42 & -  \\ 
10 x 10 & 60.08 & 50.58 & 51.74 & 49.94 & - & 64.66 & 63.00 & 63.02 & 62.08 & - & 58.50 & 60.98 & 60.13 & 67.15 & -  \\ 
15 x 15 & 70.12 & 69.03 & 62.70 & 66.16 & - & 61.63 & 60.87 & 61.43 & 61.50 & - & 52.91 & 57.60 & 54.28 & 48.52 & -  \\ 
\midrule
& \multicolumn{15}{c}{20 Labels Per Class} \\
\midrule
5 x 5 & 41.64 & 44.31 & 35.04 & 34.94 & 32.31 & \bf{79.38} & 66.07 & 63.19 & 55.52 & 63.30 & 47.96 & 47.96 & 46.86 & 46.86 & 42.84 \\ 
7 x 7 & 59.58 & 57.90 & 45.62 & 45.15 & 44.51 & 67.24 & 67.91 & 69.22 & 71.24 & 64.83 & 64.71 & 61.46 & 58.99 & 50.66 & 47.01 \\ 
10 x 10 & 71.74 & 62.92 & 52.46 & 48.63 & 52.12 & 64.83 & 67.08 & 67.51 & 63.28 & 56.96 & 62.32 & 64.79 & 56.08 & 41.01 & 33.45 \\ 
15 x 15 & 75.22 & 73.26 & 70.37 & 70.47 & 64.68 & 69.84 & 67.03 & 62.27 & 61.24 & 57.92 & 70.53 & 67.23 & 68.44 & 61.99 & 54.19 \\ 
20 x 20 & 78.84 & 72.84 & 67.94 & 68.39 & 62.13 & 72.45 & 66.94 & 64.79 & 63.90 & 59.05 & 67.90 & 64.91 & 66.58 & 64.02 & 61.72 \\ 
25 x 25 & 76.12 & 73.90 & 73.56 & 68.84 & 58.47 & 72.50 & 74.55 & 70.55 & 71.25 & 57.99 & 70.18 & 70.13 & 69.44 & 68.74 & 64.39 \\ 
\midrule
& \multicolumn{15}{c}{30 Labels Per Class} \\
\midrule
5 x 5 & 45.56 & 50.22 & 42.85 & 39.41 & 47.96 & 77.63 & 75.88 & 74.34 & 64.05 & 70.68 & 46.02 & 42.44 & 44.73 & 42.78 & 45.41 \\ 
7 x 7 & 63.64 & 66.10 & 49.11 & 48.94 & 42.03 & 71.24 & 71.52 & 67.33 & 60.88 & 59.88 & 67.34 & 68.38 & 68.38 & 66.50 & 64.08 \\ 
10 x 10 & 69.00 & 68.71 & 64.28 & 70.78 & 66.17 & 74.06 & 69.40 & 68.68 & 68.12 & 62.65 & 70.51 & 69.80 & 67.20 & 58.11 & 69.05 \\ 
15 x 15 & 75.11 & 75.00 & 69.27 & 68.43 & 63.97 & 74.74 & 74.12 & 73.50 & 67.76 & 64.61 & 73.46 & 71.06 & 69.70 & 68.78 & 63.45 \\ 
20 x 20 & 77.09 & 74.46 & 73.79 & 73.83 & 71.21 & 80.37 & 78.37 & 73.05 & 70.67 & 59.90 & 70.41 & 64.95 & 70.54 & 70.76 & 66.12 \\ 
25 x 25 & 78.43 & 75.93 & 73.11 & 72.78 & 71.06 & 80.85 & \bf{82.44} & 79.63 & 75.96 & 69.18 & 73.32 & 73.82 & 71.94 & 72.62 & 68.58 \\ 
\midrule
& \multicolumn{15}{c}{40 Labels Per Class} \\
\midrule
5 x 5 & 51.77 & 44.43 & 38.07 & 38.29 & 42.72 & 72.58 & 74.95 & 74.02 & 71.01 & 57.53 & 42.32 & 42.32 & 39.69 & 43.47 & 40.33 \\ 
7 x 7 & 58.65 & 58.76 & 57.88 & 54.29 & 59.99 & 75.33 & 70.48 & 67.49 & 67.58 & 66.83 & 61.63 & 59.75 & 58.70 & 58.97 & 60.47 \\ 
10 x 10 & 73.19 & 73.84 & 63.91 & 63.48 & 63.53 & 76.85 & 71.53 & 70.52 & 66.47 & 64.18 & 61.38 & 65.61 & 65.61 & 71.06 & 66.66 \\ 
15 x 15 & 73.36 & 74.45 & 70.35 & 70.22 & 67.02 & 68.47 & 65.82 & 63.59 & 60.34 & 60.74 & 69.26 & 65.51 & 64.82 & 67.85 & 61.64 \\ 
20 x 20 & 77.06 & 74.28 & 75.31 & 73.45 & 66.95 & 76.35 & 73.66 & 72.87 & 71.69 & 66.34 & 70.70 & 68.59 & 66.99 & 64.95 & 62.23 \\ 
25 x 25 & 76.04 & \bf{77.67} & 74.37 & 73.76 & 69.70 & 76.89 & 76.71 & 72.95 & 73.16 & 66.57 & 70.03 & 71.37 & 67.33 & 65.95 & 66.90 \\
\bottomrule
\end{tabular}
\end{table*}

\section{Generating a Labeled Phase of Flight Dataset}
\label{sec:dataset}

The National General Aviation Flight Information Database (NGAFID)~\cite{labella2022optimized,yang2022largescale} has been developed through funding by the Federal Aviation Administration as a program responsible for curating, storing, and analyzing GA flight data. As of January 2024, the NGAFID houses over 2.2 million flight hours of per second flight data recorder (FDR) data, collected from over 698 aircraft across 143 fleets and 303 individual users. While the NGAFID provides periodic releases of data, the data within it is mostly unlabeled apart from predefined parametric exceedances.

Aircraft at institutions participating in the NGAFID~\cite{yang2022largescale} have flight data recorders (FDRs) installed which typically capture per-second data from a comprehensive set of sensors. After each flight, the data cards from the FDRs are removed from the aircraft and uploaded to the NGAFID through a web interface, either individually or in bulk.  The flight data recorder captures a comprehensive set of parameters of the flight every second throughout its entire duration, spanning from taxiing to takeoff, and until taxiing after landing as shown in Figure~\ref{fig:Flight_Phases}. As many of the institutions participating in the NGAFID are flight schools, the flight data is more complex than typically found in commercial aviation, with student pilots testing multiple takeoffs and landings, as well as recovery maneuvers and other less common flight patterns.

This work utilizes a manually selected set of 6 representative flights of Cessna-172S aircraft (the most common aircraft in the NGAFID), which had data collected from their Garmin G1000 FDRs. These were selected due to having multiple examples of each phase of flight.  We utilized the NGAFID to export files for each containing 64 different sensor parameters.  These were preprocessed to remove data extraneous or confusing to the problem of the phase of flight detection (e.g., latitude and longitude) and also to remove rows of data early and late in the flight where some sensors are not yet online or providing erroneous information while they are being calibrated.

Following this, manual labeling of each row (representing each second) of flight data into respective flight phases was done, as the definitions provided by the NTSB (Table~\ref{tab:freq} presents each phase of flight as defined by the NTSB~\cite{flight_phases_ntsb}) and the data from the FDRs lack the granularity required for automated labeling.  To accurately label each flight phase, a thorough examination of altitude variations, initial and final airport altitudes, ground speed, and indicated airspeed was performed, and rows that could potentially be multiple phases of flights were further validated with input from aviation experts who are users of the NGAFID. This process is extremely time-consuming, which limits the number of files able to be labeled, and also serves as a motivation for this work. These files, labeled with per row phases of flight, serve as the basis for training and evaluating the accuracy of the SOM variations described in Section~\ref{sec:methodology}. Table~\ref{tab:row_counts} presents the total number of rows in each flight file as well as how many were labeled as each phase, highlighting the severe class imbalance of these phases.

Each of the preprocessed and labeled flight files has been stored as CSVs for easy use and has been made publicly available at National General Aviation Flight Information Database\footnote{http://ngafid.org}. They contain sensor data from the following parameters: Nearest Airport Distance (AirportDistance), Altitude Above Ground Level (AltAGL), Altitude Above Sea Level Lagged Differential (AltMSL Lag Diff), amp1, amp2, Angle of Attack (AOASimple), Calibrated Airspeed (CAS), Coordination Index, Engine 1 Cylinder Head Temperature Divergence (E1 CHT Divergence), Engine 1 Cylinder Head Temperature 1 (E1 CHT1), Engine 1 Cylinder Head Temperature 2 (E1 CHT2), Engine 1 Cylinder Head Temperature 3 (E1 CHT3), Engine 1 Cylinder Head Temperature (E1 CHT4), Engine 1 Exhaust Gas Temperature  Divergence (E1 EGT Divergence), Engine 1 Exhaust Gas Temperature 1 (E1 EGT1), Engine 1 Exhaust Gas Temperature 2 (E1 EGT2), Engine 1 Exhaust Gas Temperature 3 (E1 EGT3), Engine 1 Exhaust Gas Temperature 4 (E1 EGT4), Engine 1 Fuel Flow (E1 FFlow), Engine 1 Oil Pressure (E1 OilP), Engine 1 Oil Temperature (E1 OilT), E1 RPM, Fuel Quantity Left Tank (FQtyL), Fuel Quantity Right Tank (FQtyR), Ground Speed (GndSpd), Horizontal Limit Alert (HAL), Heading (HDG), HPLfd, Indicated Airspeed (IAS), Lateral Acceleration (LatAc), Loss of Control Index (LOC-I Index), NAV1, NAV2, Norm Acceleration (NormAc), Outside Air Temperature (OAT), Pitch, Roll, Nearest Runway Distance (RunwayDistance), Stall Index, True Airspeed (TAS), Total Fuel, Ground Track (TRK), volt1, volt2, Vertical Protection Level (VPLwas), Vertical Speed (VSpd), VSpd Calculated, and VSpdG; along with a final column for the labeled phase of flight (FlPhase).

\section{Results}

Experiments were designed to determine how well the MS-SOM methodology described in Section~\ref{sec:ms_som_method} compares to the naive approach presented in Section~\ref{sec:naive_som_method}. We evaluated MS-SOM using a varying limited number of labeled samples along with the naive SOM method using an entire labeled flight file. SOM sizes ranging from 5x5, 7x7, 10x10, 15x15 to 20x20 were evaluated for these two methodologies. Additionally, for the minimally labeled SOM we investigated extracting 5, 10, and 20 labels of each class from a single flight file, as well as extracting 5, 10, 20, 30, and 40 labels randomly selected from a set of flight files. The minimally labeled SOM methodologies were tested using 3, 5, 7, 10, and 15 nearest neighbors for inference.  As data normalization could affect the model training process and potentially greatly affect the performance of the model, all experiments were tested using three data normalization methods: min-max, standard (z-score), and robust. Each experimental setup was repeated 10 times on a MacBook Pro with an Intel Core i9 processor.

\subsection{Naive SOM}

For baseline results, a naive SOM was trained for each of the selected SOM sizes using the flight 53438 data file, and 20 labeled data points for each class (except for \emph{Take Off} which only had 16 labeled rows) were selected to measure validation accuracy of the SOM while it was training. After training the naive SOM was tested on the other 5 flight files. Table~\ref{tab:som_38} shows the validation and testing accuracy of naive SOM. The best validation and testing results for each data normalization method are marked in bold. The overall best-found validation accuracy was $85.17\%$ and the best testing accuracy was $83.35\%$, both coming from the min-max normalization method with a SOM of size 15x15.

The U-matrix (unified distance matrix) in Figure~\ref{fig:umatrix-Naive-som} is a representation of a self-organizing map (SOM) where the Euclidean distance between the codebook vectors of neighboring neurons is depicted in a density-based image.
The right side of the graph contains a bar that signifies the color density in the map matrix. The more dense the color, the more the codebook vectors are widely separated and vice versa. Thus, a light section in the matrix denotes a cluster and the dark parts can be noted as boundaries between the clusters. As seen in Figure~\ref{fig:umatrix-Naive-som}, we can see several clusters that have been created. Each dot within the scatter plots in the U-matrix cell depicts the mapping of a row for different flight phases of our flight. This shows that the phases of flight are well clustered together in different regions on the map. 

\subsection{Minimally Supervised SOM}

To investigate the limits of how little data is required to train an effective SOM for the phase of flight detection, two sets of experiments were conducted. The first set only drew labeled data points from the flight 53438 file, with 5, 10, and 20 labeled data points for each class (except for the Take Off class which only had 16 data points, all of which were used in the 20 data point experiment). To further investigate the effect of increasing the amount of labeled data points from each class, the second set of experiments used 5, 10, 20, 30, and 40 labeled data points randomly selected from flight files 53434, 53436, 53438 (these rows were then removed from the testing data for these experiments). Given these different numbers of labeled data points, we tested a wide range of different SOM sizes with different numbers of nearest neighbors used for class inference. Table~\ref{tab:train_config} presents the different configurations of SOM size and number of nearest neighbors used for different numbers of labels.

Each trained SOM was tested on all 6 flight files (with labeled data removed). Tables~\ref{tab:38_test_all} and~\ref{tab:343638_test_all} show the testing results for those two sets of experiments. The best testing accuracy for each particular number of labeled data points per class is marked in bold. For the first set of experiments, the testing accuracy of $82.99\%$ and $82.49\%$ are very close and both were found using min-max normalization methods. For the second set of experiments, the best-performing SOM had a testing accuracy of $82.44\%$. The best-performing accuracies were found to be close, showing that for flight phase detection, drawing labeled data points from one single flight file does not bring bias to training and testing performance. The two tables also show increasing the number of labels for minimally supervised SOM generally increases the model performance to a point, as the best results were found with 30 labels per class. Interestingly, after reaching that optimal performance, adding more labeled data began to decrease the overall model performance.

\subsection{Effects of Class Imbalance}

Since the flight data is class imbalanced, being able to predict the minority classes is crucial and shows the robustness of the model. Tables~\ref{tab:mini_som_38} and~\ref{tab:mini_som_343638} show the best validation and testing per-class prediction accuracy for the two sets of experiments. The column on the right shows the mean of the per-class prediction accuracy. From the tables we observe that the per-class prediction performance increases as more labeled data points are used for training and the model prediction ability is more balanced among all classes. We especially see significant validation and testing performance increases for the minority classes 1--Take off and 4--Landing.

Table~\ref{tab:som_vs_mssom} compares the per-class prediction accuracy of the best found SOMs trained with different size from the naive SOM and from each of the two sets of minimal SOM experiments. For MS-SOM models, as the SOM size increases, the ability to predicting minority classes grow significantly, and the prediction performance for each class is more balanced. While for Naive SOM models, the prediction performance for minority classes 1 and 4 actually decreases as the SOM size increases. The minimally supervised models have more balanced per-class accuracy. In fact, the SOM trained using labeled rows only from Flight 53438 significantly outperformed on the smaller classes. Given the number of labels in the minimal SOM models, the results show our proposed model is effective and robust while requiring significantly less manual effort in labeling data.


\begin{table}[!t]\centering
\caption{Flight 53438 SOM Validation and Test Per 
 Class Accuracy}\label{tab:mini_som_38}
\scriptsize
\begin{tabular}{c|cccccc|c}\toprule
Label &  0 & 1 & 2 & 3 & 4 & 5 &  Mean\\
\midrule
\multicolumn{8}{c}{Validation} \\
\midrule
5  & 80.00 & 60.00 & 40.00 & 60.00 & 80.00 & 100.00 & 70.00 \\
10  & 60.00 & 90.00 & 80.00 & 80.00 & 90.00 & 90.00 & 81.67 \\
20  & 85.00 & 93.75 & 90.00 & 95.00 & 90.00 & 95.00 & 91.46 \\ 
\midrule
\multicolumn{8}{c}{Testing} \\
\midrule
5  & 80.55 & 40.46 & 68.59 & 62.17 & 85.04 & 84.77 & 70.26 \\ 
10  & 87.45 & 45.04 & 71.65 & 77.51 & 74.02 & 84.02 & 73.28 \\ 
20  & 86.48 & 93.89 & 64.89 & 71.81 & 84.25 & 84.63 & 80.99 \\ 

\bottomrule
\end{tabular}
\end{table}


\begin{table}[!t]\centering
\caption{3 Flight SOM Validation and Test Per Class Accuracy}\label{tab:mini_som_343638}
\scriptsize
\begin{tabular}{c|cccccc|c}\toprule
Label &  0 & 1 & 2 & 3 & 4 & 5 &  Mean\\
\midrule
\multicolumn{8}{c}{Validation} \\
\midrule
5 & 80.00 & 0.00 & 80.00 & 100.00 & 0.00 & 80.00 & 56.67 \\
10 & 80.00 & 90.00 & 40.00 & 80.00 & 20.00 & 80.00 & 65.00 \\ 
20 & 80.00 & 95.00 & 85.00 & 75.00 & 90.00 & 70.00 & 82.50 \\
30 & 83.33 & 60.00 & 76.67 & 90.00 & 80.00 & 80.00 & 78.33 \\ 
40 & 82.50 & 65.00 & 70.00 & 75.00 & 77.50 & 85.00 & 75.83 \\
\midrule
\multicolumn{8}{c}{Testing} \\
\midrule
5 & 88.68 & 3.05 & 75.95 & 73.57 & 8.27 & 65.54 & 52.51 \\ 
10 & 85.19 & 29.77 & 84.72 & 53.69 & 12.60 & 67.62 & 55.60 \\ 
20 & 90.45 & 9.16 & 69.89 & 79.05 & 32.28 & 77.54 & 59.73 \\ 
30 & 75.20 & 63.36 & 71.23 & 76.24 & 38.98 & 86.77 & 68.63 \\ 
40 & 84.50 & 70.99 & 80.28 & 74.61 & 81.50 & 76.04 & 77.99 \\ 

\bottomrule
\end{tabular}
\end{table}


\begin{table}[!t]\centering
\caption{Per Class Testing Accuracy}\label{tab:som_vs_mssom}
\scriptsize
\begin{tabular}{c|cccccc|c}\toprule
 Size & 0 & 1 & 2 & 3 & 4 & 5 &  Mean\\
\midrule
\multicolumn{8}{c}{Naive SOM} \\
\midrule
5x5 & 56.93 & 35.88 & 66.34 & 61.39 & 38.58 & 87.30  & 77.97 \\
10x10 & 69.85 & 30.53 & 85.95 & 64.17 & 53.54 & 85.99 & 80.79 \\ 
15x15 & 68.49 & 16.03 & 88.03 & 77.31 & 64.17 & 87.96 &83.35 \\
20x20 & 88.68 & 0.00 & 82.64 & 74.08 & 17.32 & 80.62 & 80.92 \\
25x25 & 97.95 & 0.00 & 58.13 & 73.35 & 0.00 & 83.74 & 82.70 \\
\midrule
\multicolumn{8}{c}{MS-SOM Flight 53438} \\
\midrule
5x5   & 72.24 & 0.00 & 44.33 & 54.39 & 34.25 & 89.77 & 49.16 \\
7x7  & 85.22 & 0.00 & 62.75 & 60.54 & 48.43 & 86.04 & 57.16 \\
10x10 & 92.15 & 52.67 & 61.58 & 59.25 & 68.50 & 85.00 & 69.86 \\  
15x15 & 87.45 & 45.04 & 71.65 & 77.51 & 74.02 & 84.02 & 73.28 \\ 
20x20 & 80.23 & 73.28 & 71.23 & 60.24 & 72.83 & 85.43 & 73.87 \\ 
25x25 & 86.48 & 93.89 & 64.89 & 71.81 & 84.25 & 84.63 & 80.99 \\ 
\midrule 
\multicolumn{8}{c}{MS-SOM 3 Flights} \\
\midrule
5x5  & 97.06 & 0.00 & 51.30 & 78.52 & 0.00 & 79.00 & 50.98 \\
7x7  & 74.43 & 0.00 & 21.51 & 83.60 & 31.89 & 80.85 & 48.71 \\ 
10x10 & 88.53 & 0.76 & 77.61 & 73.01 & 43.70 & 74.96 & 59.76 \\ 
15x15 & 92.65 & 6.11 & 82.43 & 84.98 & 11.81 & 69.32 & 57.88 \\
20x20 & 82.38 & 52.67 & 75.99 & 75.34 & 65.35 & 81.24 & 72.16 \\ 
25x25 & 75.20 & 63.36 & 71.23 & 76.24 & 38.98 & 86.77 & 68.63 \\
\bottomrule
\end{tabular}
\end{table}

\section{Conclusion and Future Work}

This work investigates the use of minimally supervised self-organizing maps (MS-SOMs) a novel approach for minimally supervised learning, and applies them to the phase of flight identification, an important problem in the field of aviation. While SOMs and other machine learning methods have been used as methods for phase of flight identification, training of these methods can take significant amounts of labeled training data. Unfortunately, labeling this data is both time-consuming and potentially expensive as it can require expert feedback and input for high-quality labeled data.

In this work we utilized data extracted from the National General Aviation Flight Information Database (NGAFID), and manually labeled the phase of flight for each row in a set of 6 flight sensor data files (over 41,000 rows) from Cessna-C172S aircraft, utilizing expert input to determine the phase of flight for challenging data. This data has been made available to the community for replicability and the development of a new phase of flight identification methods.

Using this data, we investigated the limits of how little data is required to perform an effective phase of flight identification. We found that the MS-SOM strategy was able to reach similar or better levels of overall accuracy using as few as 30 labeled rows per class, compared to vanilla SOMs utilizing an entire file (3781 rows) of labeled data, and also able to significantly outperform the vanilla approach for the class labels which had low representation. This is particularly important given the high-class imbalance of this data. These results suggest that MS-SOMs are an effective approach for minimally supervised learning, and can be used as an effective phase of flight identification method.

These results also open up multiple avenues for future work. In particular, the SOMs were trained using per-row data. Utilizing multiple rows per training sample could provide additional information for a better phase of flight identification. Further, aircraft do not tend to rapidly change between phases of flight, so the results from the MS-SOM could be incorporated into a system that uses a rolling average of previous phases of flight predictions to smooth out SOM predictions, potentially increasing accuracy significantly. Lastly, the NGAFID contains data from over 15 types of aircraft, so training MS-SOMs for other aircraft in the system using less data can bring significant benefit as these methods get incorporated into the NGAFID.

\bibliographystyle{IEEEtran}
\bibliography{references}
\end{document}